\documentclass[journal]{IEEEtran}
\usepackage{graphicx}
\usepackage{booktabs}
\usepackage{subfigure}
\usepackage{amsmath}
\usepackage{algorithm}
\usepackage{algorithmic}
\usepackage{amssymb}
\usepackage{hyperref}
\usepackage{multirow}
\usepackage[table]{xcolor}
\usepackage{makecell}
\hypersetup{
	colorlinks=true,
	linkcolor=black,
	urlcolor=black,
	citecolor=black
}

\begin{document}
\title{Mask Clustering-based Annotation Engine for Large-Scale Submeter Land Cover Mapping}
\author{
    Hao Chen,
    Fang Xu,
    Tamer Saleh,
    Weifeng Hao, and
    Gui-Song Xia

    \thanks{This work is supported by the National Natural Science Foundation of China (No.~62325111, 62401406) and the Fundamental Research Funds for the Central Universities (No.~2042025kf0076). (\textit{Corresponding author: Fang Xu})

    Hao Chen is with the State Key Laboratory of Information Engineering in Surveying, Mapping and Remote Sensing (LIESMARS), Wuhan University, Wuhan 430079, China (e-mail: \texttt{chenhao20@whu.edu.cn}).

    Fang Xu, Tamer Saleh, and Gui-Song Xia are with the School of Artificial Intelligence, Wuhan University, Wuhan 430079, China (e-mail: \texttt{xufang@whu.edu.cn}; \texttt{tamersaleh@whu.edu.cn}; \texttt{guisong.xia@whu.edu.cn}).

    Weifeng Hao is with the Chinese Antarctic Center of Surveying and Mapping, Wuhan University, Wuhan 430079, China (e-mail: \texttt{haowf@whu.edu.cn}).}
}

\markboth{IEEE Transactions on Geoscience and Remote Sensing}
{Shell}

\maketitle

\begin{abstract}
Recent advances in remote sensing technology have made submeter resolution imagery increasingly accessible, offering remarkable detail for fine-grained land cover analysis. However, its full potential remains underutilized—particularly for large-scale land cover mapping—due to the lack of sufficient, high-quality annotated datasets.
Existing labels are typically derived from pre-existing products or manual annotation, which are often unreliable or prohibitively expensive, particularly given the rich visual detail and massive data volumes of submeter imagery.
Inspired by the spatial autocorrelation principle, which suggests that objects of the same class tend to co-occur with similar visual features in local neighborhoods, we propose the Mask Clustering-based Annotation Engine (MCAE), which treats semantically consistent mask groups as the minimal \textcolor{black}{annotating} units to enable efficient, simultaneous \textcolor{black}{annotation} of multiple instances. 
It significantly improves annotation efficiency by one to two orders of magnitude, while preserving label quality, semantic diversity, and spatial representativeness. 
With MCAE, we build a high-quality annotated dataset of \textit{about 14 billion} labeled pixels, referred to as HiCity-LC, which supports the generation of city-scale land cover maps across five major Chinese cities with classification accuracies above 85\%.
It is the \textit{first publicly available submeter resolution city-level land cover benchmark}, highlighting the scalability and practical utility of MCAE for large-scale, submeter resolution mapping.
The dataset is available at \url{https://github.com/chenhaocs/MCAE}.
\end{abstract}


\IEEEpeerreviewmaketitle

\section{Introduction}

\begin{figure*}[!t]
    \centering
    \includegraphics[width=.96\textwidth]{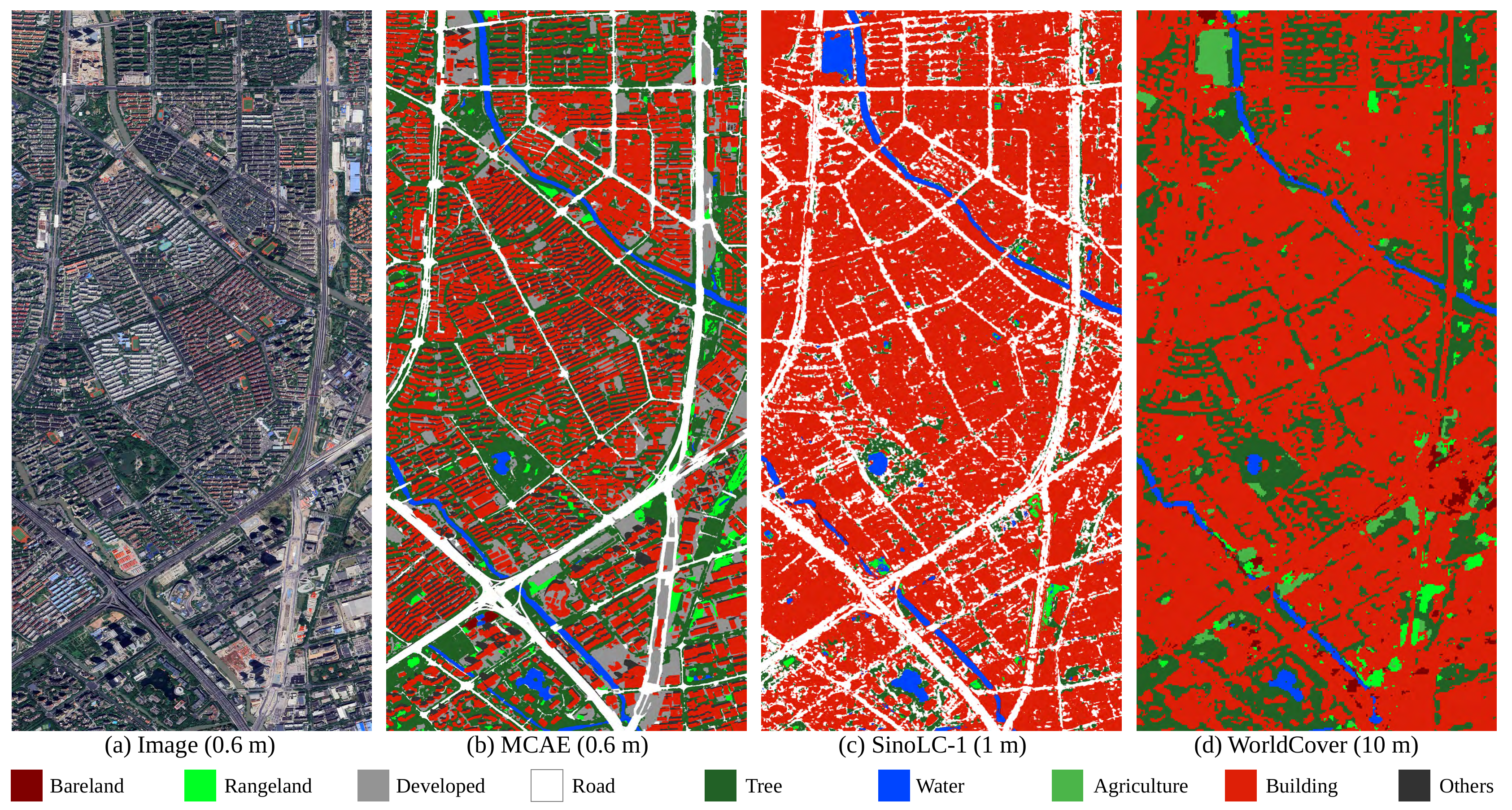}
    \caption{
        Land cover maps of the Pudong New Area, Shanghai. (a) Remote sensing imagery; (b) the result from the proposed MCAE-based method; (c) SinoLC-1~\cite{li2023sinolc}; (d) WorldCover~\cite{zanaga2021esa}. \textcolor{black}{For better visual comparison, all maps are uniformly resized for display.} For consistent visual comparison, the class systems of SinoLC-1 and ESA WorldCover have been aligned with the category system used by MCAE. Complete land cover visualizations for MCAE and SinoLC-1 are available at \url{http://pubrs.com/MCAE/} and \url{http://pubrs.com/SinoLC-1/}, respectively.
    }
    \label{fig:disp-highlight}   
    \vspace{-0.4cm}
\end{figure*}

With recent advancements in remote sensing technologies, submeter resolution satellite imagery has become increasingly accessible over extensive regions, offering rich spatial and semantic detail crucial for fine-grained surface type discrimination and boundary delineation. 
Despite these advantages, the potential of such high-resolution data for large-scale land cover mapping remains largely underexploited. 
Existing large-scale land cover products~\cite{brown2022dynamic, zanaga2021esa, venter2022global, gong2019stable} continue to rely predominantly on medium- to low-resolution sources, such as Landsat and Sentinel, whose limited spatial detail constrains the accurate identification of complex surface structures and small-scale features.
\textcolor{black}{ For instance, WorldCover, developed by the European Space Agency (ESA) using Sentinel imagery, provides global 10-meter resolution land cover maps, yet struggles to delineate narrow roads, small buildings, and fine-scale urban structures (see Fig.~\ref{fig:disp-highlight}(d)).
}

To bridge this gap, SinoLC-1~\cite{li2023sinolc,li2022breaking} \textcolor{black}{leverages high-resolution optical imagery, such as Gaofen-2 (0.8 meter), to produce a national-scale land cover map of China at 1 m spatial resolution. It adopts a weakly supervised learning strategy by transferring coarse semantic labels from existing 10-meter products (e.g., ESA WorldCover) onto fine-resolution imagery for model training, thereby alleviating the need for manual annotations on high-resolution data.}
However, its quality is fundamentally constrained by the limitations of the underlying products, which often suffer from inconsistencies in classification schemes, temporal misalignment, and inherent classification errors. These issues propagate through the refinement process, resulting in blurred boundaries and reduced mapping reliability, as shown in Fig.~\ref{fig:disp-highlight}(c).

Meanwhile, recent studies~\cite{zhang2022urbanwatch, zhang2025scale, liu2025cpvf, tian2025national, dong2021high, dong2020improving, moortgat2022deep, dong2023large} have attempted to leverage high-resolution imagery alongside finely annotated data to improve mapping quality. However, most of these efforts are confined to localized regions and lack generalizability to broader spatial scales.
A key barrier lies in the heavy reliance of current data-driven approaches on annotated datasets that are high in quality, sufficient in quantity, and spatially representative \cite{chen2024coarse, saleh2024dam}.
At submeter resolution, however, the construction of such datasets becomes increasingly impractical. \textcolor{black}{Existing submeter datasets, such as LoveDA~\cite{wang2021loveda} and OpenEarthMap~\cite{xia2023openearthmap}, remain limited in both spatial coverage and annotation density, rendering them inadequate for large-scale mapping. The core challenge lies in the intrinsic complexity of high-resolution imagery,} characterized by dense clusters of small objects and highly intricate boundary structures, which necessitates a significantly smaller minimum mapping unit (MMU)~\cite{brown2022dynamic}, thereby drastically increasing the labor intensity, time requirements, and associated costs of the annotation process.
Moreover, pronounced intra-class variability and inter-class ambiguity inherent to submeter imagery further hinder the creation of diverse and representative labeled samples, as human annotators struggle to capture the full extent of these complexities across extensive geographic areas.

Given these challenges, developing annotation methods that can substantially reduce annotating effort while preserving the diversity and representativeness of labeled data emerges as a
critical need.
Contrary to conventional approaches that regard submeter annotation as an inherently burdensome task, we propose to exploit the principle of spatial autocorrelation in geographic theory~\cite{getis2008history, xu2023harnessing} as a valuable prior to mitigate annotation costs.
It suggests that objects belonging to the same class tend to cluster within localized neighborhoods, i.e., multiple instances of the same class are likely to co-occur within a given local region of the image. 
Moreover, objects of the same class within a neighborhood often exhibit similar visual characteristics, such as color, texture, and shape. 
For example, houses in the same rural village commonly exhibit consistent architectural styles, characterized by uniform roof colors or even identical building layouts. Likewise, land parcels such as cotton fields, orchards, and aquaculture ponds often appear in spatial clusters, due to industrial-scale land use planning.

Motivated by these observations, we introduce a novel annotation paradigm that considers object clusters of the same semantic class as the minimal \textcolor{black}{annotating} unit. Under this paradigm, a single annotation operation can effectively label all instances within a cluster, substantially improving \textcolor{black}{annotation} efficiency.
Specifically, we propose the Mask Clustering-based Annotation Engine (MCAE), which features a streamlined processing pipeline comprising four key components: (1) \textit{multi-scale mask generation}, which captures object instances across varying spatial scales to account for the large variation in object sizes in high-resolution imagery; (2) \textit{mask-level feature learning}, which employs a self-supervised framework to learn semantically discriminative representations that facilitate reliable grouping of object masks; (3) \textit{hierarchical mask clustering}, which progressively aggregates masks from small to large spatial neighborhoods, ensuring both
cluster purity and completeness; and (4) \textit{iterative test set curation}, which designs an iterative spatial sampling-based label refinement strategy that builds upon, and continuously enhances, a classification model trained on prior sparse and dense annotations. 
Building upon these components, MCAE establishes a scalable and efficient annotation pipeline tailored for large-scale submeter mapping tasks. Comprehensive experiments across five representative cities show that annotations produced by MCAE enable land cover classification models to achieve overall accuracies exceeding 85\% at submeter resolution, demonstrating its capacity to provide high-quality, representative supervision for reliable and scalable land cover mapping.

\begin{figure*}[!t]
    \centering
    \includegraphics[width=.98\textwidth]{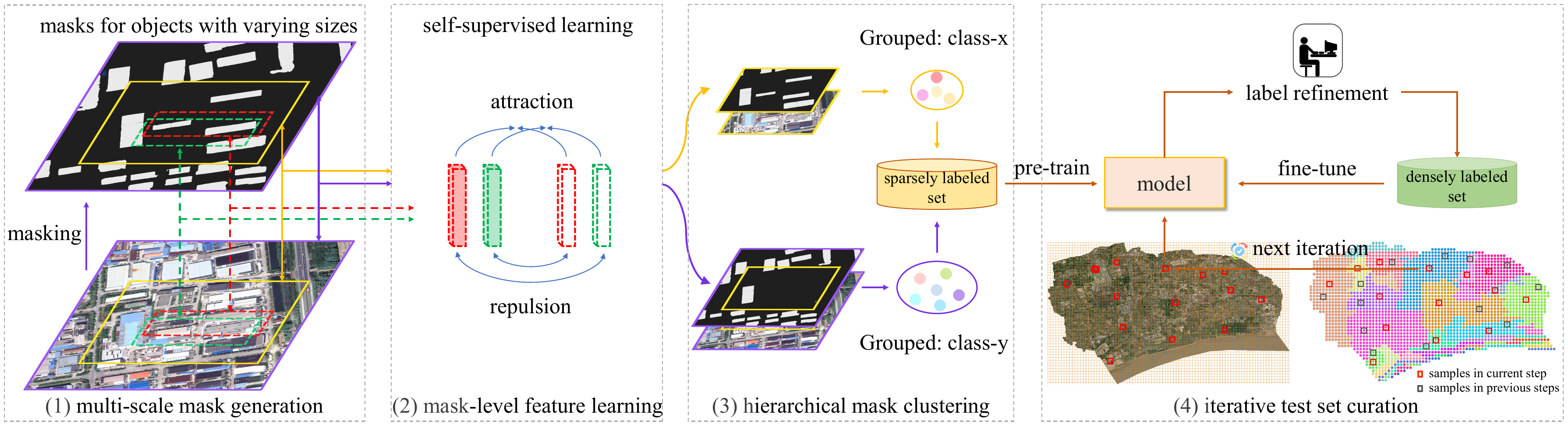}
    \caption{Overall framework of MCAE. It begins with multi-scale mask generation to capture objects of varying sizes~(1), followed by mask-level feature extraction. A self-supervised learning strategy enhances the discriminative power of these features~(2), enabling accurate clustering of semantically similar objects across spatial neighborhoods. Efficient cluster-level manual annotation yields diverse and representative labels~(3), which serve as training data for generating initial predictions. These predictions are manually refined to construct a densely annotated dataset for evaluation~(4).}
    \label{fig:overall}
    \vspace{-0.4cm}
\end{figure*}

The main contributions are summarized as follows:

\begin{itemize}
  \item We propose a novel annotation paradigm for remote sensing, which introduces mask clusters as the fundamental unit of \textcolor{black}{annotation}. 
  It enables highly scalable annotation and substantially reduces manual effort, achieving one to two orders of magnitude improvement in annotation efficiency compared to conventional pixel- or mask-based approaches.

  \item We develop MCAE, the first semi-automatic annotation framework for large-scale land cover mapping at 0.6-meter resolution.  It systematically integrates multi-scale mask generation, discriminative mask-level feature representation, the balance between cluster purity and completeness, and efficient representative test set construction.

  \item We release the first set of city-scale land cover maps at 0.6-meter resolution, along with a spatially diverse dataset spanning five major Chinese cities, termed \textbf{HiCity-LC}. It comprises 5,116 densely annotated and 50,771 sparsely annotated image-label pairs (each at $1024\times1024$ pixels), totaling over 14 billion labeled pixels.
\end{itemize}

\section{Related Works}

Recent advances in global- and national-scale land cover mapping have achieved spatial resolutions of up to 10 meters \cite{brown2022dynamic, zanaga2021esa, venter2022global, gong2019stable}, primarily leveraging open-access satellite imagery from the Sentinel and Landsat satellite series. While such resolutions are adequate for broad-scale monitoring, growing demands for highly detailed land cover information—particularly in urban environments and ecologically sensitive areas—are driving a shift toward meter- and submeter-level mapping. These high-resolution efforts, typically conducted at city or regional scales, rely heavily on high-quality annotated data. Depending on their source, such annotations can be broadly categorized into two types: existing public datasets and \textcolor{black}{human manual annotation}.

Public datasets provide a valuable foundation for training land cover classification models. Among them, OpenStreetMap (OSM) \cite{OpenStreetMap} is a crowdsourced data source that contains a large number of labels with global coverage and diverse annotation types, including scene-level, object-level, and point-level labels. It can be used for the thematic mapping of buildings, roads, and urban functional zones, or as a source of labels for a few categories in land cover mapping. However, its spatial coverage is highly uneven, with a strong bias toward urban areas, limiting its applicability in rural and natural landscapes. Datasets such as OpenEarthMap (OEM) \cite{xia2023openearthmap} and Five-Billion-Pixels (FBP) \cite{tong2023enabling} offer submeter or meter-level annotations across wide geographic regions—44 countries for OEM and nationwide coverage in China for FBP. Despite their broad scope, the annotation density within each region remains relatively sparse, constraining their utility in region-specific applications. The Chesapeake Bay Program Land Cover (CBP) \cite{chesapeake2020landcover} provides high-quality annotations over several eastern U.S. states based on 1-meter NAIP imagery \cite{usgs2020earthexplorer}. Although CBP offers reliable labels, its limited geographic extent restricts generalizability to other regions such as China and Europe. Coarse-resolution land cover products \cite{brown2022dynamic, zanaga2021esa, venter2022global} released by authoritative agencies can also serve as another source of annotation for high-resolution land cover mapping. These are treated as noisy labels, and through the design of weakly supervised learning \cite{li2022breaking, li2023sinolc} or algorithms that mine reliable labels \cite{chen2023novel}, high-spatial-resolution land cover mapping can be achieved. But this approach is often constrained by the classification system of coarse-resolution products and limited accuracy.

Manual annotation plays a critical role, especially for mapping specific regions where public datasets are unavailable or insufficient. Due to the high cost and labor requirements associated with \textcolor{black}{annotating} high-resolution imagery, the quantity of manual annotations is typically not large, and the coverage is not extensive, usually restricted to cities, sub-national regions, or countries with a small area. UrbanWatch \cite{zhang2022urbanwatch}, for instance, achieves city-level urban mapping through a combination of manual annotation and semi-supervised learning methods. By leveraging existing public datasets (e.g., OEM) and supplementing them with targeted manual annotations based on initial classification results, it is possible to reduce the annotation effort while still achieving satisfactory mapping accuracy \cite{yokoya2024submeter}.
In addition to multi-class land cover mapping, several domain-specific studies have focused on the fine-scale classification of particular land cover types—such as agricultural field parcels \cite{liu2025cpvf}, river systems \cite{moortgat2022deep}, and mangrove forests \cite{zhang2025scale}. These tasks require submeter or even ultra-high spatial resolution to meet stringent application-specific accuracy requirements. To this end, researchers frequently rely on dense manual annotation, including pixel-level or point-level labels, to achieve the desired performance.
Despite these advances, a persistent bottleneck remains: the lack of a generalizable, geographically and categorically scalable, and cost-effective framework for generating high-quality annotations on high-resolution imagery.

\section{Methodology}

The overall framework of the proposed Mask Clustering-based Annotation Engine (MCAE) is illustrated in Fig.~\ref{fig:overall}.
MCAE is designed to efficiently generate large-scale, high-quality annotations for submeter remote sensing imagery through a four-stage pipeline.
The process begins with the generation of precise multi-scale masks, enabling the accurate capture of objects with diverse spatial scales. 
These masks provide the foundation for mask-level feature extraction, which encodes rich semantic information at the object level to support the clustering of semantically similar instances.
To enhance feature discriminability, a self-supervised learning strategy is employed, encouraging masks of the same class to exhibit similar representations while pushing apart those of different classes.
By progressively clustering objects across neighborhoods with increasing spatial scales, manual annotation is efficiently performed at the cluster level, resulting in a large number of spatially sparse mask annotations that are both diverse and discriminative. 
The sparse annotations generated through cluster-level \textcolor{black}{annotation} are then used as training data to produce initial pixel-wise label predictions for spatially sampled images. These predictions are manually verified and corrected, constructing a densely annotated dataset for evaluation.

\begin{figure}[!ht]
    \centering
    \includegraphics[width=1.0\columnwidth]{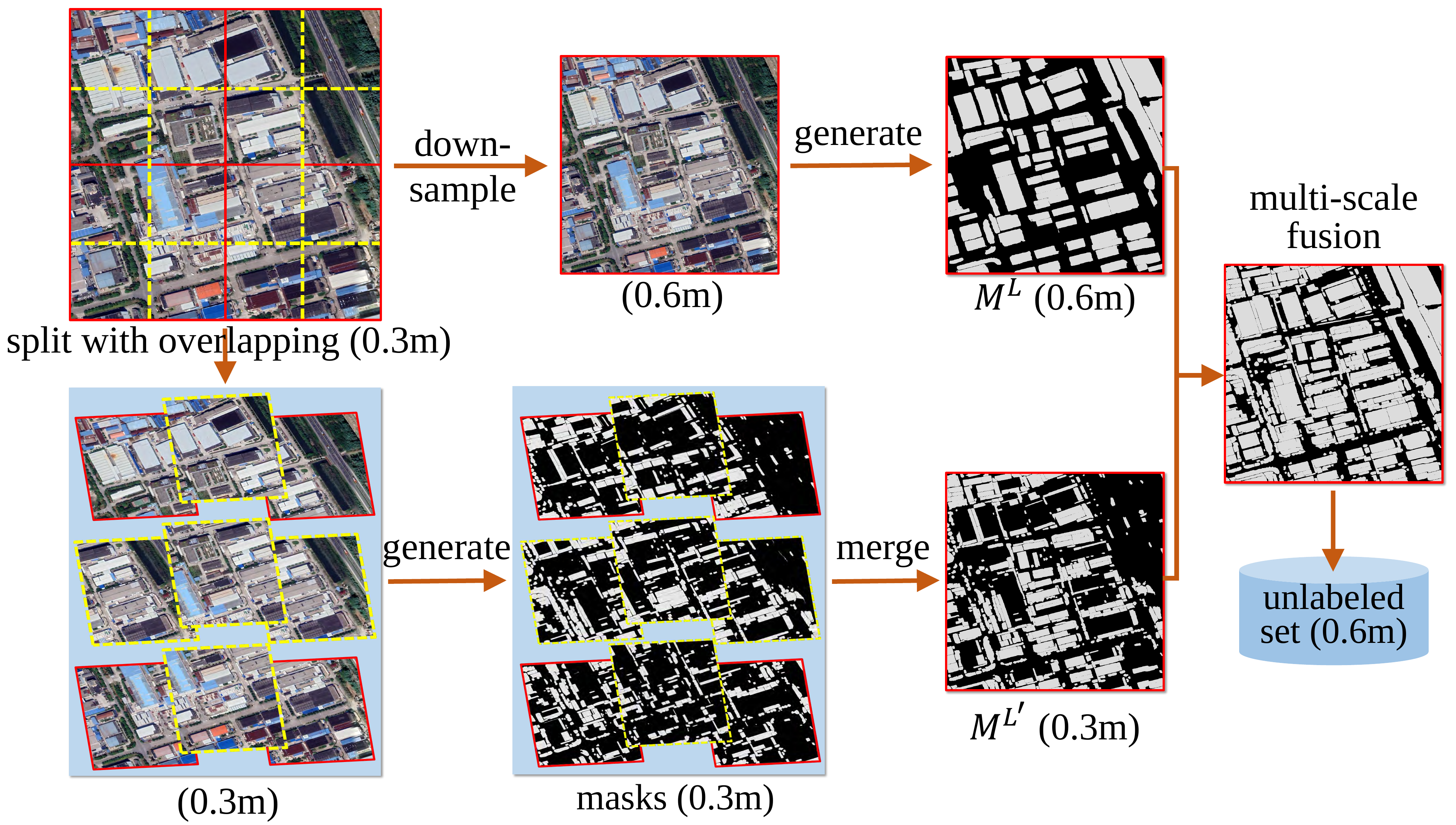}
    \caption{Detail of multi-scale mask generation.}
    \label{fig:mask-fusion}
    \vspace{-0.4cm}
\end{figure}
\subsection{Multi-scale mask generation} \label{Multi-scale-mask-gen}

Generating accurate object masks is a fundamental prerequisite for land cover semantic annotation. Recently, a series of powerful foundation segmentation models—such as the Segment Anything Model (SAM)~\cite{kirillov2023segment}—have emerged as promising tools for rapid and high-quality mask generation. However, their performance in remote sensing scenarios, where objects exhibit substantial variability in size, remains limited. In particular, the effectiveness of these models often degrades when applied to high-resolution remote sensing imagery, primarily due to a pronounced sensitivity to object scale. When input images are processed at a coarse scale, small objects are likely to be overlooked or only partially segmented, resulting in incomplete or missing object masks. In contrast, at finer scales, large objects may extend beyond the effective receptive field of the model or be misclassified as background.
To address this, we adopt a multi-scale strategy that captures object boundaries across diverse spatial extents, as shown in Fig.~\ref{fig:mask-fusion}.

Specifically, given an original image with a spatial resolution of 0.3 meters, object masks are generated at both 0.3-meter and 0.6-meter scales, and subsequently fused to obtain a set of refined object masks.
For the 0.3-meter scale, the image is partitioned into overlapping tiles with a 50\% overlap ratio, following the tiling scheme illustrated in Fig.~\ref{fig:mask-fusion}, to preserve object continuity across tile boundaries and enhance segmentation completeness for small-scale objects.
Each tile is independently processed by the off-the-shelf SAM \cite{kirillov2023segment}, \textcolor{black}{a foundation segmentation model trained on over 11 million images and 1.1 billion masks, which demonstrates strong zero-shot generalization across diverse segmentation tasks, to generate object masks. Specifically, we adopt the ViT-L variant\footnote{\url{https://huggingface.co/facebook/sam-vit-large}} as the segmentation backbone due to its superior performance in capturing fine-grained object boundaries.}
For any two overlapping tiles, if their predicted masks in the overlapping region are consistent—i.e., they represent the same object with matching boundaries—the mask is retained; otherwise, both conflicting masks are discarded to avoid semantic ambiguity and boundary artifacts caused by inconsistent segmentation across overlapping regions.
The resulting set of refined masks obtained at the 0.3-meter scale is denoted as $M^{L'}$.
For the 0.6-meter scale, the original image is first downsampled to the corresponding resolution and then processed by SAM to generate a complementary set of coarse-resolution masks, denoted as $M^{L}$.

To fully leverage the complementary strengths of the two scales—namely, the boundary precision of fine-scale masks and the contextual completeness of coarse-scale masks—we perform a multi-scale mask fusion between $M^{L'}$ and $M^{L}$. 
The fusion procedure follows the principle of favoring a finer-scale mask to preserve object purity. In particular, while decomposing a large object into smaller, well-defined components typically does not degrade classification accuracy, preserving an entire coarse-scale object may result in boundary leakage, introducing pixels from adjacent categories and thus compromising label quality.
To implement this fusion, all non-overlapping masks from $M^{L'}$ and $M^{L}$ are directly retained. For overlapping masks,
each mask in $M^{L'}$ is compared with its corresponding counterpart in $M^{L}$. If a finer-scale mask is entirely contained within a coarser one, the latter is divided into two regions: one that exactly matches the finer mask, and \textcolor{black}{the other} representing the remaining area. In cases of partial overlap, a three-part decomposition is applied, producing the overlapping region and the non-overlapping segments from both masks.

\begin{figure}[!ht]
    \centering
    \includegraphics[width=1.0\columnwidth]{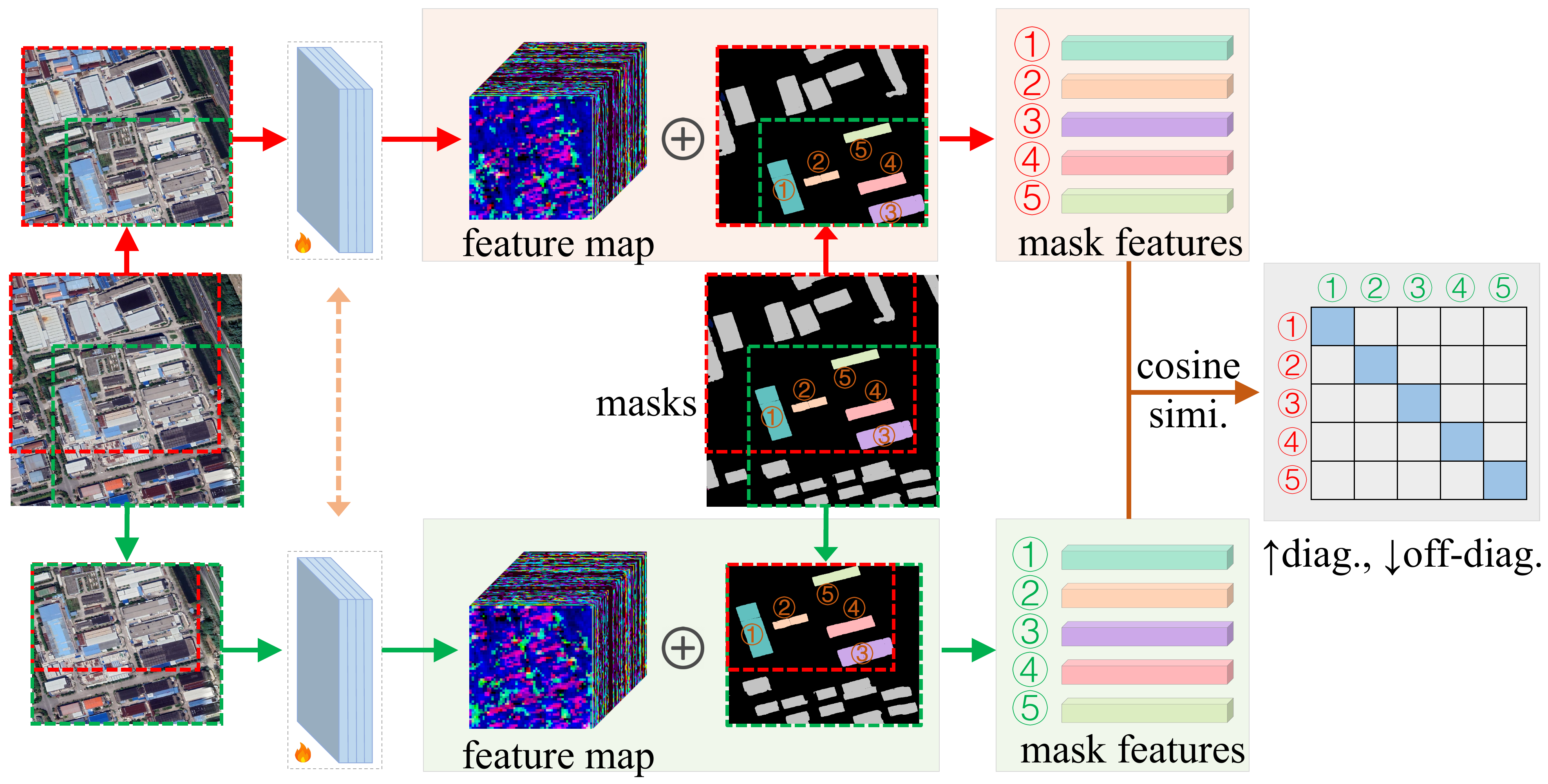}
    \caption{Detail of mask-level feature learning.}
    \label{fig:obj-contrast}
    \vspace{-0.4cm}
\end{figure}

\subsection{Mask-level feature learning} \label{Mask-level-feature-learning}

\begin{figure*}[!ht]
    \centering
    \includegraphics[width=0.85\textwidth]{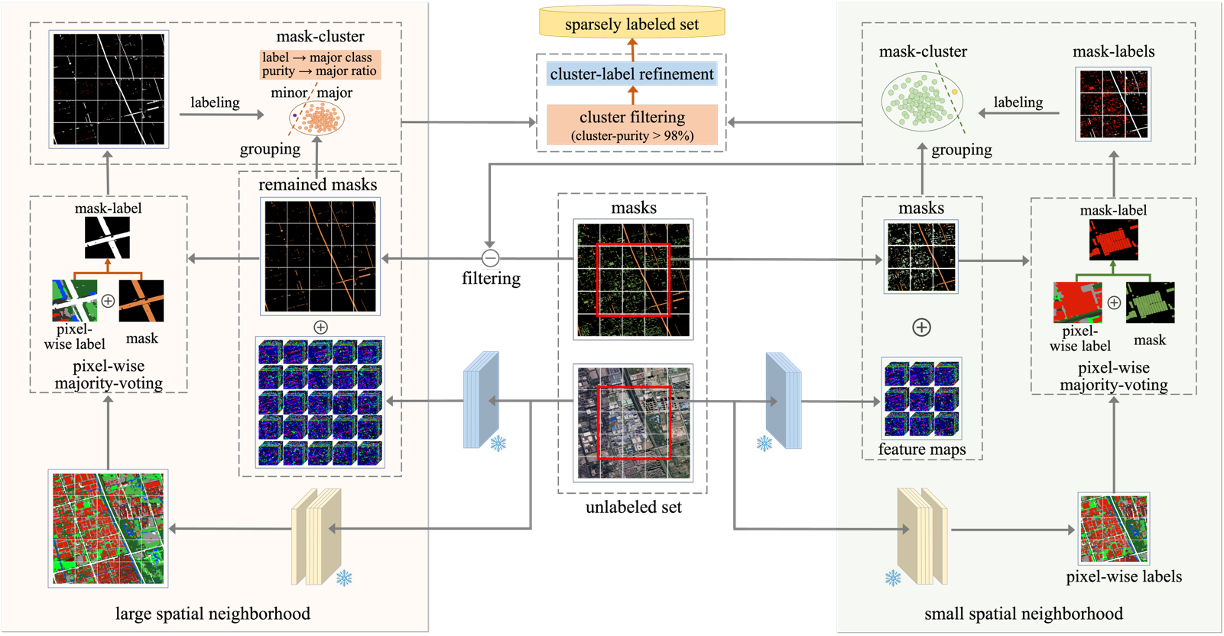}
\caption{Detail of hierarchical mask clustering.}
\vspace{-0.4cm}
\label{fig:cluster-ann}
\end{figure*}

Clustering object masks into semantically meaningful groups relies on high-quality mask-level features. Although pre-trained models and previously generated object masks together can be used to extract such features, they are typically derived from pixel-level features obtained via pre-trained weights, rather than from genuine mask-level features. As a result, these approximations may fail to capture fine-grained inter-class distinctions, hindering the effectiveness of clustering. To address this limitation, a dedicated feature learning approach is required to extract object-centric embeddings, with the goal of improving intra-class compactness and inter-class separability, especially within local spatial neighborhoods.
In this paper, we design a self-supervised learning approach to extract enhanced mask-level features by encouraging semantic consistency across instances of the same mask and separation between different masks, as shown in Fig.~\ref{fig:obj-contrast}. 

We first generate two overlapping crops from an input tile and extract their corresponding feature maps using a Swin Transformer backbone~\cite{liu2021swin}. For each mask present in the overlapping region of the two crops, the associated features are used to construct a similarity matrix. Features from the same mask are encouraged to be similar, while those from different masks are expected to be distinct. A self-supervised loss is then formulated to enforce the contrastive relationship, thereby facilitating the learning of discriminative and semantically consistent mask-level representations.

\begin{figure*}[!t]
    \centering
    \includegraphics[width=0.85\textwidth]{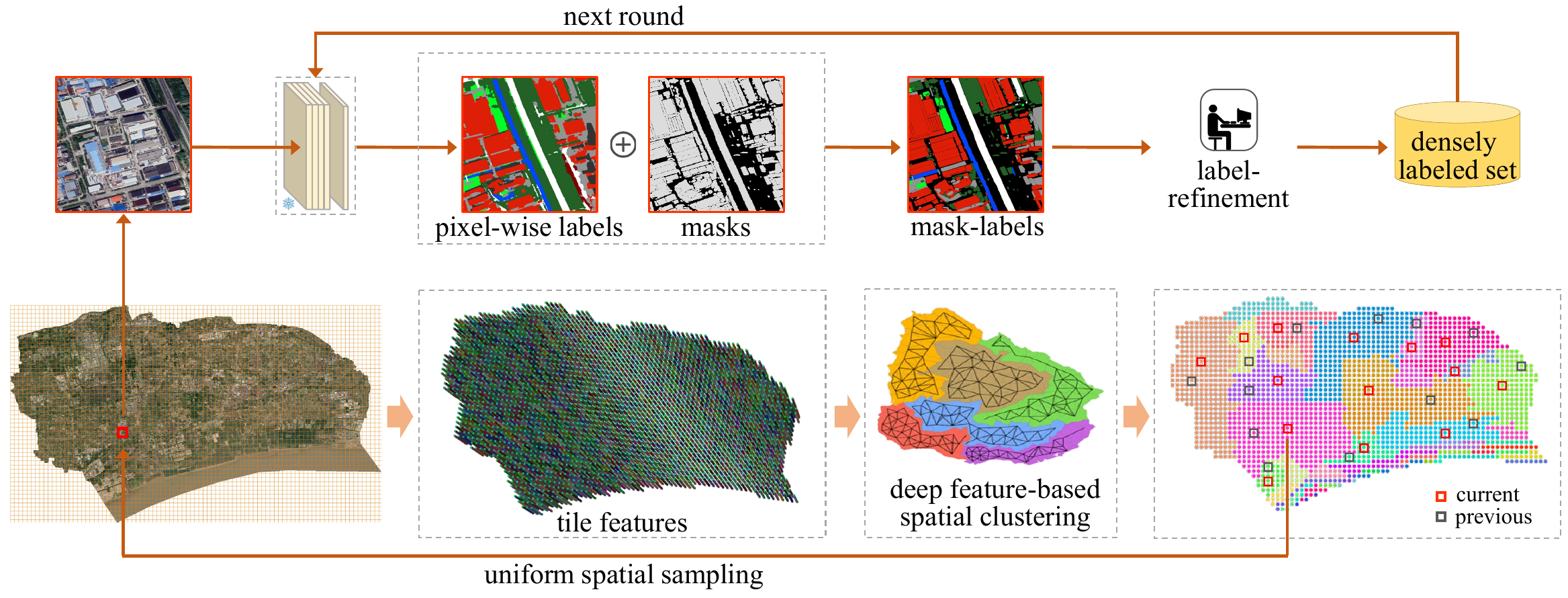}
    \caption{Detail of iterative test set curation. }
    \label{fig:sampling-refine}
    \vspace{-0.4cm}
\end{figure*}

\subsection{Hierarchical mask clustering} \label{Hierarchical-mask-clustering}

The spatial aggregation of semantically similar masks enables the simultaneous \textcolor{black}{annotation} of entire mask clusters, thereby significantly reducing annotation workload. However, different categories of masks exhibit diverse spatial distribution patterns, with some being densely concentrated and others more sparsely distributed. 
To enhance annotation efficiency under the premise of cluster purity, it is beneficial to prioritize larger clusters, as they typically contain more semantically consistent masks and reduce the frequency of manual \textcolor{black}{annotation}.
However, it may overlook sparsely distributed masks, thereby decreasing overall cluster completeness.
Conversely, incorporating smaller clusters can enhance completeness but often contain limited information due to their small size, leading to unstable feature representations. 
Such instability may blur cluster boundaries and introduce noise, ultimately degrading cluster purity and increasing the annotation burden. 
Therefore, achieving a balance between cluster purity and completeness is a crucial challenge in designing effective mask clustering strategies.

Building upon the high-quality object masks and discriminative mask-level features obtained from the previous stages, we develop a hierarchical mask clustering strategy to achieve efficient and scalable label acquisition, as illustrated in Fig.~\ref{fig:cluster-ann}. 
Concretely, clustering is first conducted within a relatively small neighborhood.
The DBSCAN algorithm~\cite{martin1996dbscan}, a density-based clustering method that automatically determines the number of clusters and is robust to noise, is adopted.
To ensure that the resulting clusters are suitable for unified annotation, we incorporate a semantic consistency-based filtering step.
Specifically, a semantic segmentation model—the \textit{UPerNet-Swin-Small} variant\footnote{\url{https://huggingface.co/openmmlab/upernet-swin-small}} pre-trained on the OpenEarthMap dataset~\cite{xia2023openearthmap}—is used to generate pixel-level category predictions.
Each mask is assigned a semantic label via majority voting over the predicted pixel classes within its spatial extent. Cluster purity is then measured as the proportion of masks sharing the dominant class within each cluster. Only clusters with sufficiently high purity are retained to ensure reliable unified annotation, with the dominant class serving as a reference label to assist manual annotation.

To improve annotation completeness, the remaining masks not labeled in the small-neighborhood clustering are further processed within a larger spatial neighborhood. While small-scale clustering is effective for identifying and \textcolor{black}{annotating} densely distributed objects, it often overlooks semantically consistent instances that are sparsely distributed across the image. By expanding the clustering scope, these spatially dispersed but semantically related masks can be aggregated into coherent clusters, enabling unified annotation. This hierarchical strategy ensures both the semantic purity of individual clusters and broader coverage of diverse object instances, thereby enhancing the quality and representativeness of the final annotations. In this paper, the small neighborhood corresponds to a 3×3 tile window, and the large neighborhood to a 5×5 tile window.

\subsection{Iterative test set curation} \label{Test-set-curation}

Reliable evaluation of models trained on the sparsely labeled data generated through the mask clustering-based annotation requires the availability of densely labeled test datasets that comprehensively represent the entire target area in terms of class diversity, spatial distribution, and annotation quality and quantity. However, existing datasets and products rarely meet these requirements, and manually constructing such datasets from scratch entails significant cost and time investment.

While the earlier steps primarily generate sparse annotations sufficient for training models with strong generalization capabilities, such annotations are inadequate for comprehensive performance evaluation. 
To thoroughly assess a model’s predictive performance across the entire spatial extent and all semantic categories—while controlling annotation costs—it is essential to construct a densely labeled test set that is both spatially distributed and semantically diverse. 
To this end, we introduce an iterative test set curation strategy that combines model predictions with spatial sampling to systematically identify and prioritize the informative regions for detailed annotation, thereby enabling efficient, reliable, and scalable model evaluation, as shown in Fig.~\ref{fig:sampling-refine}.

For each large-area image covering a specific region, the curation process begins by applying a pooling operation over the feature map of each tile to obtain a compact vector representation. These representations are then used to perform spatial partitioning. Specifically, we adopt the SKATER algorithm~\cite{assunccao2006efficient}, which constructs a minimum spanning tree based on feature similarity while enforcing spatial adjacency constraints. It ensures that the resulting partitions are not only semantically coherent but also spatially contiguous, making them suitable for downstream sampling and annotation.

To ensure that the test set is both spatially representative and semantically diverse, a fixed number of tiles are uniformly sampled from each partition. As dense and accurate annotations are required for evaluation, directly \textcolor{black}{annotating} the test set would be costly. To reduce annotation overhead, we utilize a semantic segmentation model—specifically, the \textit{UPerNet-Swin-Small} variant, consistent with the architecture used in Sec.~\ref{Hierarchical-mask-clustering}—which has been trained on the previously generated sparse annotations to produce initial pixel-wise predictions.
Given that the model has been trained on spatially distributed and semantically diverse sparse annotations, its predictions are expected to exhibit a relatively low error rate.
The predicted labels are overlaid with corresponding object masks to form preliminary annotation maps, which are then manually refined to correct classification errors and boundary inaccuracies.

To further reduce \textcolor{black}{annotation} costs while improving annotation quality, we adopt a multi-round refinement strategy. After the first round of sampling and refinement, a small number of densely labeled samples are obtained and used, together with the original sparse annotations, to retrain the model for better regional adaptation. The updated model guides subsequent rounds of sampling and refinement, progressively yielding a high-quality, densely annotated test set for reliable model evaluation.

\subsection{Evaluation Metrics}

We evaluate the experimental results using four standard semantic segmentation metrics: Overall Accuracy (OA), mean F1-score (mF1), mean Intersection over Union (mIoU), and User’s Accuracy (UA). Specifically:

\begin{itemize}
    \item  \textbf{Overall Accuracy (OA)} quantifies the proportion of correctly classified pixels across the entire dataset:
    \[
    \text{OA} = \frac{\sum_{i=1}^{K} TP_i}{\sum_{i=1}^{K} (TP_i + FP_i + FN_i + TN_i)}
    \]
    where \( TP_i \), \( FP_i \), \( FN_i \), and \( TN_i \) denote the number of true positives, false positives, false negatives, and true negatives for class \( i \), respectively, and \( K \) is the total number of classes.

    \item \textbf{mean F1-score (mF1)} mF1 (mean F1-score) reflects the average harmonic mean of precision and recall across all classes, providing a balanced measure of classification quality, especially under class imbalance:
    \[
    \text{mF1} = \frac{1}{K} \sum_{i=1}^{K} \text{F1}_i, 
    \text{F1}_i 
    = \frac{2TP_i}{2TP_i + FP_i + FN_i},
    \]

    \item \textbf{mean Intersection over Union (mIoU)} evaluates the average ratio between the intersection and union of predicted and ground-truth segments across all classes:
    \[
    \text{mIoU} = \frac{1}{K} \sum_{i=1}^{K} \text{IoU}_i, 
    \text{IoU}_i = \frac{TP_i}{TP_i + FP_i + FN_i},
    \]

    \item \textbf{User’s Accuracy (UA)} measures the reliability of predictions for each class, defined as the proportion of correctly predicted pixels among all pixels assigned to that class:
    \[
    \text{UA}_i = \frac{TP_i}{TP_i + FP_i}
    \]
\end{itemize}

\section{Experimental results} \label{Experiment}

\begin{figure}[!b]
    \centering
    \includegraphics[width=1.0\columnwidth]{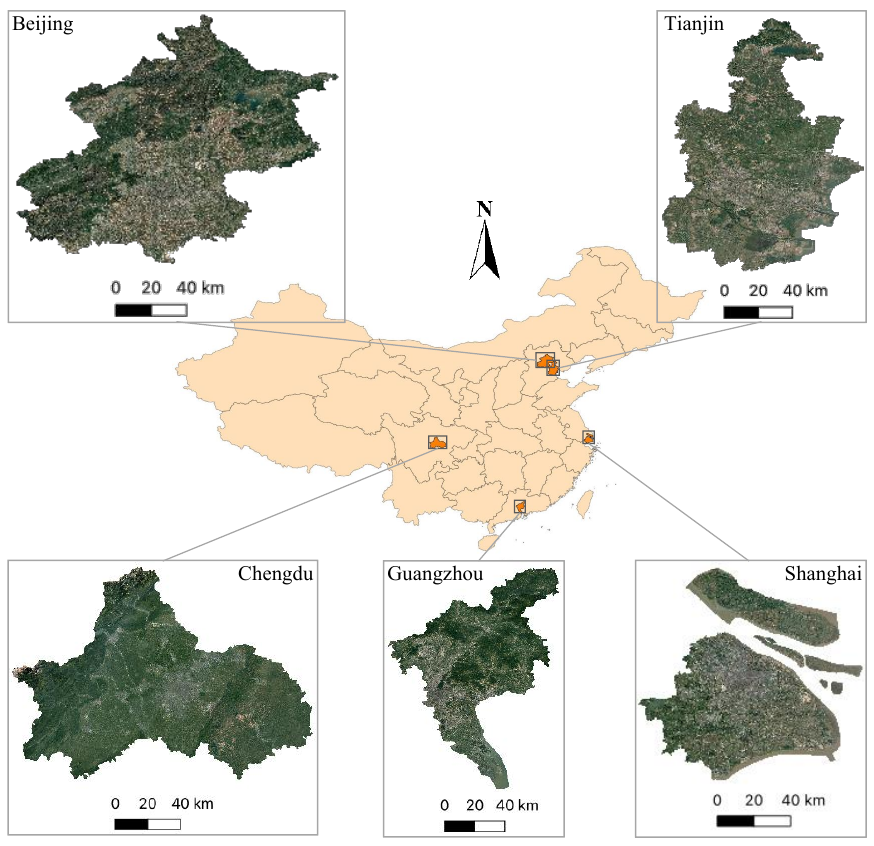}
    \caption{The study areas selected—Beijing, Tianjin, Shanghai, Guangzhou, and Chengdu—are representative of diverse land cover characteristics across northern, eastern, southern, and western regions of China.}
    \label{fig:study-area}
    \vspace{-0.4cm}
\end{figure}

\subsection{Study area and data}

To evaluate the effectiveness of the proposed data annotation method, experiments are conducted on satellite imagery from five representative Chinese cities—Shanghai, Chengdu, Guangzhou, Beijing, and Tianjin—as shown in Fig.~\ref{fig:study-area}. The satellite imagery is sourced from Google Earth with a spatial resolution of 0.3 meters. \textcolor{black}{Detailed characteristics of the study areas, including their areas, data volumes, and image dimensions, are summarized in Tab.~\ref{tab:stats-study-areas}.} Given the massive volume of city-scale imagery, each city is subdivided by administrative units (districts or counties) during preprocessing and annotation, with those of small area excluded.
The classification system follows that of OpenEarthMap~\cite{xia2023openearthmap}, which includes eight land cover categories: bareland, rangeland, developed space, road, tree, water, agricultural land, and building.
Based on the above experimental regions and leveraging the proposed MCAE framework, we construct a large-scale land cover dataset named \textbf{HiCity-LC}. It comprises 55,887 image–label pairs with a spatial resolution of 0.6 meters and a size of 1024×1024 pixels, totaling approximately 14 billion labeled pixels. In particular, 5,116 images are labeled with dense mask annotations (approximately 3.74 billion pixels), while the remaining 50,771 images contain sparse mask annotations (approximately 10.27 billion pixels).

\begin{table}[!t]
  \centering
  \caption{Overview of Selected Study Areas and Data Characteristics}
  \label{tab:stats-study-areas}
  \resizebox{\linewidth}{!}{
  \begin{tabular}{lrrc}
    \toprule
    \textbf{City Names} & \textbf{Area ($\text{km}^2$)} & \textbf{Data Volumes (GB)} & \textbf{Dimensions (px)} \\
    \midrule
    Shanghai   & 6,341  & 647   & 518{,}656 × 446{,}464 \\
    Chengdu    & 12,132 & 996   & 710{,}144 × 502{,}016 \\
    Guangzhou  & 7,434  & 607   & 410{,}880 × 528{,}896 \\
    Beijing    & 16,411 & 1,287 & 780{,}032 × 603{,}392 \\
    Tianjin    & 11,760 & 895   & 506{,}368 × 632{,}832 \\
    \midrule
    \textbf{Total} & 54,078 & 4,432 & -- \\
    \bottomrule
  \end{tabular}}
\end{table}

\begin{table}[!t]
  \centering
  \caption{Statistics of mask clusters and object masks generated by MCAE across five cities.}
  \setlength{\tabcolsep}{2pt}
  \renewcommand{\arraystretch}{1.15}
  \resizebox{\linewidth}{!}{
    \begin{tabular}{llrrr}
    \toprule
    \textbf{City} & \textbf{District/County} & \textbf{Area (km$^2$)} & \textbf{\#Clusters} & \textbf{\#Masks} \\
    \midrule
    \multirow{3}{*}{Beijing} 
      & Daxing     & 1,036.3 & 681 & 23,525 \\
      & Changping  & 1,343.5 & 518 & 25,695 \\
      & Miyun      & 2,229.4 & 521 & 13,176 \\
    \midrule
    \multirow{3}{*}{Tianjin} 
      & Binhai     & 2,270.0 & 1,246 & 44,411 \\
      & Jizhou     & 1,589.3 & 462   & 31,874 \\
      & Wuqing     & 1,574.0 & 493   & 47,981 \\
    \midrule
    \multirow{3}{*}{Shanghai} 
      & Pudong     & 1,210.0 & 936 & 85,646 \\
      & Fengxian   & 733.4   & 290 & 10,419 \\
      & Qingpu     & 675.1   & 412 & 23,017 \\
    \midrule
    \multirow{3}{*}{Chengdu} 
      & Pengzhou   & 1,421.8 & 514 & 64,441 \\
      & Qionglai   & 1,377.0 & 404 & 46,832 \\
      & Chongzhou  & 1,089.0 & 544 & 71,751 \\
    \midrule
    \multirow{3}{*}{Guangzhou} 
      & Conghua    & 1,985.0 & 349 & 23,385 \\
      & Zengcheng  & 1,616.5 & 434 & 15,885 \\
      & Huadu      & 970.0   & 253 & 11,474 \\
    \midrule
    \multicolumn{2}{l}{\textbf{Total}} & 21,120.3 & 8,057 & 539,512 \\
    \midrule
    \multicolumn{2}{l}{\textbf{Average masks per cluster}} & & \multicolumn{2}{c}{67.0} \\
    \bottomrule
    \end{tabular}
  }
  \label{tab:5city-mask-stat}%
  \vspace{-0.4cm}
\end{table}%

\begin{table*}[!t]
  \centering
  \caption{Comparison of land-cover semantic segmentation results across MCAE and baseline annotation methods (\textit{pixel-based} and \textit{mask-based}) under a fixed annotation budget in five representative cities, evaluated by IoU, UA, F1, and OA.}
  \setlength{\tabcolsep}{2pt}
  \renewcommand{\arraystretch}{1.15}
  \resizebox{0.96\linewidth}{!}{
  \begin{tabular}{lccc|ccc|ccc|ccc|ccc}
    \toprule
    \multirow{2}{*}{Class} 
    & \multicolumn{3}{c|}{Beijing} & \multicolumn{3}{c|}{Tianjin} & \multicolumn{3}{c|}{Shanghai} 
    & \multicolumn{3}{c|}{Chengdu} & \multicolumn{3}{c}{Guangzhou} \\
    \cmidrule(lr){2-16}
    & Mask & Pixel & MCAE & Mask & Pixel & MCAE & Mask & Pixel & MCAE & Mask & Pixel & MCAE & Mask & Pixel & MCAE \\
    \rowcolor{gray!15}
    \multicolumn{16}{c}{\textbf{IoU}} \\
    \midrule
    Bareland & 11.25 & 4.66 & 45.77 & 5.79 & 1.48 & 29.27 & 3.82 & 1.98 & 17.82 & 9.77 & 3.55 & 36.78 & 3.17 & 2.67 & 22.17 \\
    Rangeland & 14.26 & 8.43 & 63.76 & 13.03 & 4.29 & 55.29 & 13.63 & 3.79 & 59.31 & 15.05 & 5.79 & 63.10 & 12.46 & 4.77 & 55.75 \\
    Developed & 12.79 & 5.47 & 61.96 & 12.73 & 5.12 & 51.09 & 12.15 & 5.30 & 54.23 & 13.42 & 8.26 & 57.55 & 10.84 & 5.69 & 55.42 \\
    Road & 16.52 & 7.57 & 71.62 & 11.93 & 6.71 & 61.23 & 16.91 & 7.34 & 72.88 & 16.12 & 9.18 & 74.94 & 13.26 & 7.53 & 66.24 \\
    Tree & 18.59 & 7.04 & 93.57 & 18.19 & 6.55 & 82.33 & 15.95 & 6.46 & 83.53 & 19.51 & 6.83 & 92.54 & 17.51 & 8.15 & 93.89 \\
    Water & 20.08 & 11.37 & 87.09 & 16.54 & 7.96 & 81.87 & 14.94 & 5.89 & 71.71 & 16.13 & 9.32 & 88.98 & 16.06 & 6.01 & 81.63 \\
    Agriculture & 17.48 & 6.49 & 79.12 & 16.65 & 7.35 & 87.71 & 18.17 & 6.77 & 75.14 & 18.34 & 9.29 & 86.23 & 13.30 & 5.46 & 76.55 \\
    Building & 19.30 & 8.68 & 84.00 & 17.16 & 7.93 & 85.55 & 17.76 & 7.19 & 77.59 & 19.45 & 7.42 & 85.66 & 16.10 & 7.57 & 81.01 \\
    Others & 16.26 & 6.94 & 61.62 & 11.40 & 6.87 & 64.10 & 13.55 & 5.19 & 52.44 & 14.70 & 6.64 & 68.13 & 13.76 & 5.02 & 55.08 \\
    \midrule
    Mean  & 16.28 & 7.41  & 72.06 & 13.71 & 6.03  & 66.49 & 14.10 & 5.55  & 62.74 & 15.83 & 7.36  & 72.66 & 12.94 & 5.87  & 65.30 \\
    % \midrule
    \rowcolor{gray!15}
    \multicolumn{16}{c}{\textbf{UA}} \\
    \midrule
    \multicolumn{1}{l}{Bareland  } & 12.35 & 5.92  & 56.09 & 12.97 & 5.96  & 58.52 & 12.46 & 4.08  & 61.32 & 10.22 & 6.62  & 44.21 & 11.84 & 3.92  & 57.55 \\
    \multicolumn{1}{l}{Rangeland  } & 14.45 & 10.97 & 75.61 & 13.07 & 4.51  & 67.29 & 21.83 & 7.62  & 83.22 & 15.37 & 9.41  & 76.00 & 13.79 & 8.60  & 70.51 \\
    \multicolumn{1}{l}{Developed  } & 19.87 & 5.48  & 78.09 & 17.42 & 5.41  & 68.88 & 20.74 & 8.03  & 75.98 & 14.24 & 8.27  & 74.29 & 20.82 & 6.96  & 75.69 \\
    \multicolumn{1}{l}{Road    } & 22.74 & 8.89  & 83.78 & 19.45 & 9.52  & 76.84 & 18.16 & 8.75  & 87.75 & 20.23 & 12.53 & 86.13 & 13.66 & 9.84  & 79.35 \\
    \multicolumn{1}{l}{Tree    } & 20.43 & 11.14 & 96.43 & 24.52 & 13.56 & 91.87 & 17.42 & 11.93 & 86.98 & 26.41 & 11.11 & 96.76 & 20.43 & 11.81 & 97.24 \\
    \multicolumn{1}{l}{Water    } & 21.68 & 12.72 & 88.57 & 18.57 & 11.12 & 84.32 & 21.00 & 12.27 & 74.05 & 22.37 & 9.47  & 92.00 & 23.26 & 7.52  & 83.14 \\
    \multicolumn{1}{l}{Agriculture } & 18.46 & 12.73 & 91.88 & 24.77 & 9.90  & 94.37 & 18.78 & 10.04 & 92.20 & 22.06 & 13.39 & 94.15 & 23.31 & 11.56 & 89.07 \\
    \multicolumn{1}{l}{Building  } & 21.42 & 12.85 & 89.40 & 23.86 & 11.27 & 90.45 & 18.00 & 9.08  & 82.14 & 21.23 & 7.46  & 90.36 & 16.16 & 10.79 & 85.08 \\
    \multicolumn{1}{l}{Others   } & 20.74 & 11.56 & 88.96 & 19.86 & 14.22 & 87.57 & 15.09 & 8.59  & 84.13 & 19.21 & 7.07  & 80.13 & 21.85 & 8.81  & 84.05 \\
    \midrule
    Mean  & 19.13 & 10.25 & 83.20 & 19.39 & 9.50  & 80.01 & 18.16 & 8.93  & 80.86 & 19.04 & 9.48  & 81.56 & 18.35 & 8.87  & 80.19 \\
    % \midrule
    \rowcolor{gray!15}
    \multicolumn{16}{c}{\textbf{F1}} \\
    \midrule
    \multicolumn{1}{l}{Bareland  } & 14.32 & 6.77  & 62.80 & 7.19  & 5.45  & 45.28 & 14.17 & 5.84  & 19.78 & 10.93 & 4.97  & 53.78 & 9.11  & 4.11  & 36.29 \\
    \multicolumn{1}{l}{Rangeland  } & 18.95 & 10.48 & 77.87 & 18.01 & 7.87  & 71.21 & 16.31 & 9.88  & 74.46 & 15.13 & 7.29  & 77.38 & 14.86 & 7.64  & 71.59 \\
    \multicolumn{1}{l}{Developed  } & 13.14 & 11.61 & 76.51 & 14.21 & 9.48  & 67.63 & 17.33 & 6.21  & 70.32 & 17.59 & 11.86 & 73.05 & 11.88 & 10.37 & 71.32 \\
    \multicolumn{1}{l}{Road    } & 23.04 & 13.55 & 83.46 & 21.50 & 8.54  & 75.95 & 18.40 & 7.40  & 84.31 & 18.82 & 10.03 & 85.68 & 22.21 & 11.32 & 79.69 \\
    \multicolumn{1}{l}{Tree    } & 26.90 & 10.51 & 96.68 & 21.64 & 9.79  & 90.31 & 23.23 & 12.36 & 91.03 & 25.42 & 10.94 & 96.13 & 23.07 & 11.18 & 96.85 \\
    \multicolumn{1}{l}{Water    } & 20.84 & 12.57 & 93.10 & 22.73 & 11.81 & 90.03 & 22.45 & 11.07 & 83.53 & 18.71 & 10.64 & 94.17 & 21.02 & 8.73  & 89.89 \\
    \multicolumn{1}{l}{Agriculture } & 24.00 & 14.17 & 88.34 & 21.41 & 9.20  & 93.45 & 18.54 & 11.30 & 85.81 & 20.44 & 14.05 & 92.60 & 24.28 & 8.83  & 86.72 \\
    \multicolumn{1}{l}{Building  } & 20.58 & 14.33 & 91.30 & 26.06 & 10.21 & 92.21 & 24.17 & 10.24 & 87.38 & 24.14 & 11.42 & 92.28 & 22.68 & 9.90  & 89.51 \\
    \multicolumn{1}{l}{Others   } & 16.88 & 10.89 & 76.25 & 18.34 & 9.04  & 78.12 & 15.90 & 8.37  & 68.80 & 21.17 & 12.03 & 81.04 & 18.82 & 8.35  & 71.03 \\
    \midrule
    Mean  & 19.85 & 11.65 & 82.92 & 19.01 & 9.04  & 78.24 & 18.94 & 9.18  & 73.94 & 19.15 & 10.36 & 82.90 & 18.66 & 8.94  & 76.99 \\
    % \midrule
    \rowcolor{gray!15}
    \multicolumn{16}{c}{\textbf{OA}} \\
    \midrule
    /    & 27.62 & 16.16 & 92.38 & 24.33 & 12.87 & 87.80 & 25.11 & 15.91 & 85.12 & 28.25 & 15.18 & 91.93 & 26.24 & 13.84 & 89.23 \\
    \bottomrule
  \end{tabular}
  }
  \label{tab:5city-peer-comparison}
  \vspace{-0.4cm}
\end{table*}

\begin{table*}[!t]
  \centering
  \caption{Comparison of land-cover semantic segmentation results between MCAE and SinoLC-1 \cite{li2023sinolc} in five representative cities, evaluated by IoU, UA, F1, and OA.}
    \setlength{\tabcolsep}{2pt}
    \renewcommand{\arraystretch}{1.15}
    \resizebox{0.96\linewidth}{!}{
    \begin{tabular}{lrr|rr|rr|rr|rr}
    \toprule
    \multirow{2}[3]{*}{Class} & \multicolumn{2}{c|}{Beijing} & \multicolumn{2}{c|}{Tianjin} & \multicolumn{2}{c|}{Shanghai} & \multicolumn{2}{c|}{Chengdu} & \multicolumn{2}{c}{Guangzhou} \\
\cmidrule{2-11}          & \multicolumn{1}{c}{SinoLC-1} & \multicolumn{1}{c|}{MCAE} & \multicolumn{1}{c}{SinoLC-1} & \multicolumn{1}{c|}{MCAE} & \multicolumn{1}{c}{SinoLC-1} & \multicolumn{1}{c|}{MCAE} & \multicolumn{1}{c}{SinoLC-1} & \multicolumn{1}{c|}{MCAE} & \multicolumn{1}{c}{SinoLC-1} & \multicolumn{1}{c}{MCAE} \\
    \rowcolor{gray!15}
    \multicolumn{11}{c}{\textbf{IoU}} \\
    \midrule    
    Bareland   & 2.10  & 51.06 & 2.15  & 55.35 & 1.23  & 40.40 & 3.61  & 44.63 & 9.63  & 59.84 \\
    Rangeland   & 4.74  & 59.93 & 0.01  & 60.99 & 0.06  & 58.58 & 2.68  & 61.63 & 0.02  & 54.81 \\
    Road     & 13.10 & 71.01 & 9.89  & 67.27 & 11.56 & 77.80 & 19.67 & 78.82 & 14.29 & 71.31 \\
    Tree     & 63.23 & 92.70 & 11.07 & 83.12 & 20.87 & 81.96 & 69.82 & 91.34 & 73.21 & 93.71 \\
    Water     & 62.16 & 90.52 & 62.93 & 92.57 & 35.00 & 88.53 & 3.75  & 90.54 & 56.72 & 96.22 \\
    Agriculture  & 25.04 & 81.52 & 57.17 & 91.99 & 33.80 & 88.89 & 35.97 & 86.08 & 20.17 & 83.02 \\
    Building   & 28.79 & 86.62 & 41.17 & 88.23 & 38.11 & 76.25 & 39.17 & 87.16 & 45.96 & 87.03 \\
    \midrule
    Mean  & 28.45 & 76.19 & 26.34 & 77.07 & 20.09 & 73.20 & 24.95 & 77.17 & 31.43 & 77.99 \\
    \rowcolor{gray!15}
    \multicolumn{11}{c}{\textbf{UA}} \\
    \midrule 
    Bareland   & 2.12  & 71.84 & 2.27  & 68.56 & 3.59  & 60.89 & 17.89 & 58.99 & 33.44 & 76.24 \\
    Rangeland   & 11.73 & 83.97 & 0.01  & 84.28 & 0.06  & 87.80 & 4.21  & 88.52 & 0.02  & 80.81 \\
    Road     & 25.12 & 87.57 & 13.19 & 83.62 & 28.88 & 90.30 & 34.51 & 90.20 & 27.45 & 85.34 \\
    Tree     & 66.78 & 93.72 & 11.30 & 87.22 & 23.15 & 83.54 & 77.01 & 92.47 & 83.76 & 95.02 \\
    Water     & 73.58 & 97.14 & 72.35 & 96.03 & 39.45 & 95.27 & 3.82  & 96.10 & 60.54 & 98.06 \\
    Agriculture  & 62.27 & 88.20 & 86.08 & 94.54 & 76.27 & 92.04 & 68.54 & 91.07 & 37.80 & 88.76 \\
    Building   & 82.83 & 88.95 & 85.98 & 90.43 & 67.06 & 82.30 & 59.72 & 89.71 & 80.93 & 90.45 \\
    \midrule
    Mean  & 46.35 & 87.34 & 38.74 & 86.38 & 34.07 & 84.59 & 37.96 & 86.72 & 46.28 & 87.81 \\
    \rowcolor{gray!15}
    \multicolumn{11}{c}{\textbf{F1}} \\
    \midrule 
    Bareland   & 4.09  & 67.60 & 4.22  & 71.26 & 2.43  & 57.55 & 6.98  & 61.72 & 17.57 & 74.87 \\
    Rangeland   & 9.04  & 74.95 & 0.02  & 75.77 & 0.12  & 73.88 & 5.22  & 76.26 & 0.03  & 70.81 \\
    Road     & 23.16 & 83.05 & 18.00 & 80.43 & 20.73 & 87.51 & 32.87 & 88.15 & 25.01 & 83.25 \\
    Tree     & 77.47 & 96.21 & 19.93 & 90.78 & 34.54 & 90.09 & 82.23 & 95.47 & 84.53 & 96.75 \\
    Water     & 76.66 & 95.02 & 77.24 & 96.14 & 51.85 & 93.92 & 7.23  & 95.04 & 72.38 & 98.07 \\
    Agriculture  & 40.05 & 89.82 & 72.75 & 95.83 & 50.53 & 94.12 & 52.90 & 92.52 & 33.57 & 90.72 \\
    Building   & 44.71 & 92.83 & 58.33 & 93.75 & 55.18 & 86.52 & 56.29 & 93.14 & 62.97 & 93.06 \\
    \midrule
    Mean  & 39.31 & 85.64 & 35.78 & 86.28 & 30.77 & 83.37 & 34.82 & 86.04 & 42.29 & 86.79 \\
    \rowcolor{gray!15}
    \multicolumn{11}{c}{\textbf{OA}} \\
    \midrule 
    /     & 62.77 & 92.21 & 61.65 & 91.79 & 44.68 & 89.04 & 64.09 & 91.38 & 69.19 & 93.78 \\
    \bottomrule
    \end{tabular}}%
  \label{tab:5city-SinoLC-comparison}
  \vspace{-0.4cm}
\end{table*}%

\subsection{Comparison with other annotation methods}

To evaluate the practical utility of our proposed annotation approach, MCAE, we compare it against two representative baselines: the \textit{pixel-based} method and the \textit{mask-based} method, both of which are implemented in widely used annotation platforms, such as Labelme\footnote{\url{https://github.com/wkentaro/labelme}} and CVAT\footnote{\url{https://github.com/cvat-ai/cvat}}.
The \textit{pixel-based} method requires users to manually select boundary pixels for each object to delineate its contour, thereby completing the annotation.
In contrast, the \textit{mask-based} method integrates models such as SAM into the annotation interface, allowing users to generate object masks by clicking a small number of point prompts.
To ensure a fair comparison, we establish an idealized upper bound for both methods. Specifically, we assume that the \textit{pixel-based} method requires a minimum of four boundary clicks per object, resulting in an annotation cost of 4 per object. For the \textit{mask-based} method, we assume that a single point prompt is sufficient to generate an accurate mask, yielding a cost of 1 per object.
Note that the estimate does not account for additional time spent on SAM inference or possible failures in mask generation.
MCAE, by contrast, leverages object clusters as the annotation unit—allowing a single annotation action to label dozens of spatially coherent instances simultaneously.
As summarized in Tab.~\ref{tab:5city-mask-stat}, MCAE yields a dramatic reduction in manual annotation workload, achieving an estimated cost of just $1/67$ per object.
It underscores its substantial advantage in minimizing human effort while ensuring large-scale, high-quality annotation coverage.

\begin{figure*}[!t]
    \centering
    \includegraphics[width=0.9\textwidth]{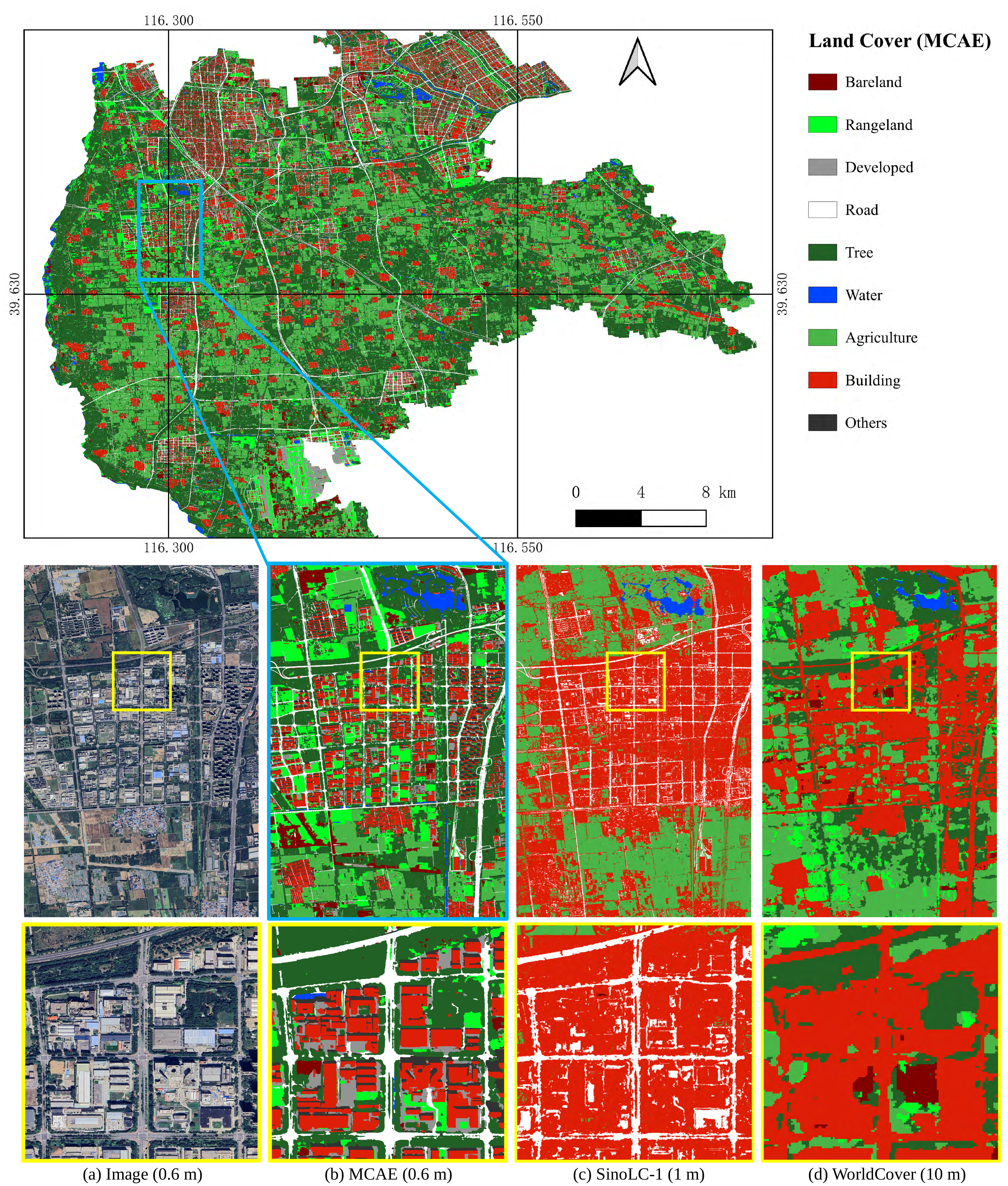}
    \caption{
      Comparison of land cover maps in the Daxing District of Beijing. (a) Remote sensing image; (b)–(d) land cover maps provided by (b) the proposed MCAE-based method, (c) SinoLC-1~\cite{li2023sinolc}, and (d)WorldCover ~\cite{zanaga2021esa}, respectively.
    }
    \label{fig:disp-daxing}   
    \vspace{-0.4cm}
\end{figure*}

\begin{table*}[!t]
  \centering
  \caption{Comparison of IoU performance using \textit{Sparse} versus \textit{Sparse + Dense} mask annotations across five representative districts.}
  \setlength{\tabcolsep}{2pt}
  \renewcommand{\arraystretch}{1.15}
  \resizebox{0.98\linewidth}{!}{
  \begin{tabular}{lcccccccccc}
    \toprule
    \multirow{2}{*}{Class} & \multicolumn{2}{c}{Beijing-Daxing} & \multicolumn{2}{c}{Tianjin-Binhai} & \multicolumn{2}{c}{Shanghai-Pudong} & \multicolumn{2}{c}{Chengdu-Pengzhou} & \multicolumn{2}{c}{Guangzhou-Conghua} \\
    \cmidrule(lr){2-3} \cmidrule(lr){4-5} \cmidrule(lr){6-7} \cmidrule(lr){8-9} \cmidrule(lr){10-11}
    & Sparse & Sparse+Dense & Sparse & Sparse+Dense & Sparse & Sparse+Dense & Sparse & Sparse+Dense & Sparse & Sparse+Dense \\
    \midrule
    Bareland        & 28.45 & 33.97 & 34.34 & 36.35 & 25.03 & 29.65 & 49.39 & 57.88 & 53.26 & 57.31 \\
    Rangeland       & 61.41 & 63.04 & 54.60 & 57.08 & 67.69 & 77.55 & 40.07 & 45.05 & 29.79 & 30.39 \\
    Developed       & 66.50 & 74.38 & 50.03 & 55.38 & 55.27 & 61.26 & 56.14 & 58.86 & 43.18 & 46.69 \\
    Road            & 78.35 & 88.51 & 62.88 & 74.54 & 66.57 & 78.85 & 67.30 & 77.17 & 52.63 & 61.28 \\
    Tree            & 91.76 & 92.00 & 60.72 & 71.80 & 84.38 & 87.87 & 80.59 & 82.58 & 90.62 & 95.95 \\
    Water           & 87.98 & 95.89 & 81.55 & 95.43 & 61.49 & 69.82 & 55.12 & 61.88 & 89.60 & 95.47 \\
    Agriculture     & 81.38 & 87.56 & 68.70 & 74.01 & 74.50 & 75.13 & 82.36 & 83.44 & 75.69 & 83.28 \\
    Building        & 86.67 & 89.81 & 69.94 & 72.61 & 82.98 & 90.39 & 84.94 & 85.96 & 65.13 & 71.32 \\
    Others          & 66.88 & 72.23 & 43.91 & 46.78 & 58.16 & 67.24 & 67.86 & 69.74 & 47.42 & 47.83 \\
    \midrule
    Mean IoU        & 72.15 & 77.49 & 58.52 & 64.89 & 61.79 & 70.86 & 66.75 & 69.17 & 61.93 & 65.50 \\
    OA              & 85.86 & 90.99 & 79.71 & 85.95 & 79.72 & 88.32 & 88.11 & 91.10 & 89.11 & 92.21 \\
    \bottomrule
  \end{tabular}}
  \label{tab:5county-mask-pixel-comparison}
\vspace{-0.4cm} 
\end{table*}

\begin{table}[!t]
  \centering
  \caption{Ablation study on multi-scale mask generation in Shanghai-Fengxian.}
  \setlength{\tabcolsep}{2pt}
  \renewcommand{\arraystretch}{1.15}
  \resizebox{\linewidth}{!}{
  \begin{tabular}{lccc}
    \toprule
    Class & 0.3-meter & 0.6-meter & Multi-scale~(Ours) \\
    \midrule
    Bareland     & 18.76 & 24.87 & 29.27 \\
    Rangeland    & 43.07 & 49.61 & 55.29 \\
    Developed    & 32.71 & 36.14 & 51.09 \\
    Road         & 59.58 & 40.99 & 61.23 \\
    Tree         & 60.33 & 76.51 & 82.33 \\
    Water        & 50.14 & 75.64 & 81.87 \\
    Agriculture  & 66.37 & 71.35 & 87.71 \\
    Building     & 70.63 & 55.52 & 85.55 \\
    Others       & 48.83 & 54.60 & 64.10 \\
    \midrule
    Mean IoU & 50.05 & 53.91 & 66.49 \\
    OA & 74.23 & 78.11 & 87.80 \\
    \midrule
    \#Masks & 912,315 & 401,064 & 1,027,335 \\
    Avg. Size (pixels) & 218$\pm 122$ & 4,243$\pm 1995$ & 1,807$\pm 1373$ \\
    \bottomrule
  \end{tabular}}%
  \label{tab:ablation-multiscale-mask-gen}
\vspace{-0.4cm} 
\end{table}

To further evaluate the advantages of MCAE under practical annotation constraints, we compare the impact of annotation quantity—under a fixed annotation budget—on model performance across different annotation methods. We first construct the training and test sets using annotations generated by MCAE. During the test set construction stage, two rounds of sampling are conducted for each district or county, with 100 samples selected in each round to form a dense and representative test set. All sparse mask annotations across the entire target area—excluding those incorporated into the test set—are aggregated to form a comprehensive training set.
We then estimate the number of annotations that the two baseline methods could generate under the same annotation cost as MCAE. Since MCAE automatically selects objects with attention to spatial uniformity and representativeness, while the baselines rely on manual selection, we enforce spatially uniform sampling for the baselines to ensure a fair comparison.
Each set of annotations is subsequently used to train a semantic segmentation model—specifically, the \textit{UPerNet-Swin-Small} variant—and the corresponding performance is reported in Tab.~\ref{tab:5city-peer-comparison}.
Across the five cities, MCAE consistently outperforms both \textit{pixel-based} and \textit{mask-based} baselines in terms of all four evaluation metrics, achieving overall accuracies of 92.38\%, 87.80\%, 85.12\%, 91.93\%, and 89.23\% in Beijing, Tianjin, Shanghai, Chengdu, and Guangzhou, respectively.
This superior performance is primarily attributed to the significantly larger volume of high-quality annotations produced by MCAE under an equivalent annotation budget. 
In contrast, the baseline methods yield considerably fewer annotations and rely on uniform sampling, which often results in suboptimal sample selection, lacking in semantic diversity and spatial representativeness—factors that further hinder their ability to train high-quality models.

\subsection{Comparison with SinoLC-1}

To our knowledge, SinoLC-1 \cite{li2023sinolc} remains the only publicly available land cover product providing high-resolution mapping at a national scale across China. It offers a 1-meter resolution classification map generated using a model trained on coarse-resolution products and limited manual annotations. Among existing products, SinoLC-1 is the most comparable to our MCAE-based results in terms of spatial resolution (0.6-meter). Therefore, we conduct both quantitative and visual comparisons to evaluate the performance of our method against SinoLC-1.

\begin{table}[!t]
  \centering
  \caption{Statistics of generated clusters and masks without mask-level self-supervised learning. }
  \setlength{\tabcolsep}{2pt}
  \renewcommand{\arraystretch}{1.15}
  % Numbers in parentheses indicate reductions relative to Tab.~\ref{tab:5city-mask-stat}.}
\resizebox{\linewidth}{!}{
  \begin{tabular}{l l r r}
    \toprule
    \textbf{City} & \textbf{District/County} & \textbf{\#Clusters} & \textbf{\#Masks } \\
    \midrule
    \multirow{3}{*}{Beijing}
      & Daxing     & 410 (\textcolor{black}{$-$271}) & 16,114 (\textcolor{black}{$-$7,411}) \\
      & Changping  & 323 (\textcolor{black}{$-$195}) & 15,311 (\textcolor{black}{$-$10,384}) \\
      & Miyun      & 327 (\textcolor{black}{$-$194}) & 7,291 (\textcolor{black}{$-$5,885}) \\
    \multirow{3}{*}{Tianjin}
      & Binhai     & 761 (\textcolor{black}{$-$485}) & 26,633 (\textcolor{black}{$-$17,778}) \\
      & Jizhou     & 293 (\textcolor{black}{$-$169}) & 19,527 (\textcolor{black}{$-$12,347}) \\
      & Wuqing     & 312 (\textcolor{black}{$-$181}) & 29,209 (\textcolor{black}{$-$18,772}) \\
    \multirow{3}{*}{Shanghai}
      & Pudong     & 572 (\textcolor{black}{$-$364}) & 49,416 (\textcolor{black}{$-$36,230}) \\
      & Fengxian   & 185 (\textcolor{black}{$-$105}) & 6,937 (\textcolor{black}{$-$3,482}) \\
      & Qingpu     & 274 (\textcolor{black}{$-$138}) & 13,438 (\textcolor{black}{$-$9,579}) \\
    \multirow{3}{*}{Chengdu}
      & Pengzhou   & 326 (\textcolor{black}{$-$188}) & 39,567 (\textcolor{black}{$-$24,874}) \\
      & Qionglai   & 262 (\textcolor{black}{$-$142}) & 25,869 (\textcolor{black}{$-$20,963}) \\
      & Chongzhou  & 354 (\textcolor{black}{$-$190}) & 39,037 (\textcolor{black}{$-$32,714}) \\
    \multirow{3}{*}{Guangzhou}
      & Conghua    & 215 (\textcolor{black}{$-$134}) & 13,098 (\textcolor{black}{$-$10,287}) \\
      & Zengcheng  & 261 (\textcolor{black}{$-$173}) & 10,075 (\textcolor{black}{$-$5,810}) \\
      & Huadu      & 151 (\textcolor{black}{$-$102}) & 6,326 (\textcolor{black}{$-$5,148}) \\
    \midrule
    \multicolumn{2}{c}{\textbf{Total}} 
      & \textbf{5,026} (\textcolor{black}{$-$3,031}) & \textbf{317,848} (\textcolor{black}{$-$221,664}) \\
    \bottomrule
  \end{tabular}}%
  \label{tab:ablation-stat-mask-level-self-sup}
\vspace{-0.4cm} 
\end{table}

Prior to quantitative comparison, two preprocessing steps were implemented: (1) spatial resampling of SinoLC-1 to 0.6-meter resolution for pixel-level alignment, and (2) harmonization of classification schemes. As the legend systems between SinoLC-1 and our results are not perfectly compatible, we restrict the evaluation to seven common classes: bareland, rangeland, road, tree, water, agriculture, and building.
Tab.~\ref{tab:5city-SinoLC-comparison} presents the comparative analysis across five representative cities using IoU, UA, F1, and OA metrics. The MCAE-based method demonstrates superiority across all metrics, which we attribute to two key factors: 1) SinoLC-1's reliance on noisy labels derived from existing coarse-resolution products during model training, 2) our framework's capability to generate high-quality annotations cost-effectively at scale through the proposed MCAE approach.

\begin{table*}[!t]
  \centering
  \caption{Comparison between HiCity-LC and representative high-resolution land-cover semantic segmentation datasets.}
  \label{tab:comp-lc-dataset}
  \resizebox{0.95\linewidth}{!}{
    \begin{tabular}{lcrrrrr}
    \toprule
    \textbf{Name} & \textbf{Year} & \textbf{Spatial Resolution (m)} & \textbf{Image Dimensions (px)} & \textbf{\#Images} & \textbf{Area ($\text{km}^2$)} & \textbf{\#Pixels (10\textsuperscript{9})} \\
    \midrule
    DeepGlobe \cite{demir2018deepglobe} & 2018 & 0.5     & 2{,}448 × 2{,}448 & 1{,}146  & 1{,}717   & 6.8 \\
    SkyScapes \cite{azimi2019skyscapes} & 2019 & 0.13    & 5{,}616 × 3{,}744 & 16       & 5.7       & -- \\
    GID \cite{tong2020land}             & 2020 & 4       & 6{,}800 × 7{,}200 & 150      & 50{,}000  & 5.0 \\
    LoveDA \cite{wang2021loveda}        & 2021 & 0.3     & 1{,}024 × 1{,}024 & 5{,}987   & 536       & 6.0 \\
    Five-Billion-Pixels \cite{tong2023enabling} & 2023 & 4 & 6{,}800 × 7{,}200 & 150      & 50{,}000  & 5.0 \\
    OpenEarthMap \cite{xia2023openearthmap}     & 2023 & 0.25–0.5 & 1{,}024 × 1{,}024 & 5{,}000   & 799       & 4.9 \\
    \midrule
    SOTA (SAMRS \cite{wang2023samrs})   & \multirow{3}{*}{2023} & 0.3–     & 1{,}024 × 1{,}024 & 17{,}480  & --        & -- \\
    SIOR (SAMRS)                        &                          & 0.5–30  & 800 × 800        & 23{,}463  & --        & -- \\
    FAST (SAMRS)                        &                          & 0.3–0.8 & 600 × 600        & 64{,}147  & --        & -- \\
    \midrule
    HiCity-LC (ours)                   & 2025 & 0.3–0.6  & 1{,}024 × 1{,}024 & 55{,}887  & 54{,}078  & 14.0 \\
    \bottomrule
    \end{tabular}
  }

  \vspace{2pt}
  {\footnotesize
    \parbox{0.95\linewidth}{
    * SAMRS is a large-scale remote sensing segmentation dataset generated by applying SAM \cite{kirillov2023segment} to existing object detection datasets (i.e., DOTA \cite{ding2021dota}, DIOR \cite{li2020object}, and FAIR1M \cite{sun2022fair1m}). It is not specifically designed for land cover mapping applications.
    }
  }
\vspace{-0.2cm}
\end{table*}

Fig.~\ref{fig:disp-daxing} provides visual comparisons for MCAE, SinoLC-1, and ESA WorldCover \cite{zanaga2021esa}. To facilitate a consistent visual comparison, the classification schemes of SinoLC-1 and ESA WorldCover have been harmonized with that of the proposed MCAE method.
Fig.~\ref{fig:disp-daxing} (b) and (f) illustrate the land cover classification results based on annotations generated by the proposed MCAE framework.
It can be observed that large-scale land cover types such as urban buildings, major roads, and woodlands are accurately recognized, while fine-grained structures, including roadside trees, pedestrian walkways within parks, and greenbelts in residential areas, are also effectively distinguished. 
It is largely attributed to the multi-scale mask generation strategy, which enhances the coverage of training samples across objects of varying spatial scales, thereby enabling the model to achieve precise segmentation of both coarse and fine spatial structures.
Moreover, although the training data are inherently sparse, the model is capable of producing dense and reliable classification results, demonstrating the strong generalization ability enabled by the MCAE-generated annotations.
In comparison, as shown in Figure~\ref{fig:disp-daxing} (c) and (d), the SinoLC-1 and ESA WorldCover exhibit deficiencies in both classification detail and spatial completeness.

\subsection{\textit{Sparse} vs. \textit{Sparse + Dense} Annotations} 
\label{Comparison-of-sparse-and-dense}

The training set generated by MCAE is inherently sparse. A key concern is whether such sparse annotations possess sufficient representativeness and diversity to support effective model training. To address this, we conduct a comparative experiment to evaluate the performance of models trained with sparse annotations alone (\textit{Sparse}) versus those trained with both sparse and dense annotations (\textit{Sparse + Dense}). Specifically, for each of the five cities, we select a relatively large district or county as the test region, while the remaining areas are used as training data.

Table~\ref{tab:5county-mask-pixel-comparison} reports the mIoU and OA results across eight land cover categories for Daxing, Binhai, Pudong, Pengzhou, and Conghua.
While models trained with \textit{Sparse + Dense} annotations achieve slightly higher performance overall, those trained solely on \textit{Sparse} annotations perform comparably across most categories. 
It suggests that the sparse annotations generated by MCAE exhibit strong spatial representativeness and semantic diversity, thereby providing a robust and cost-efficient foundation for training high-performing models.

\subsection{Ablation study}

\noindent{\textbf{Multi-scale mask generation}}.
To generate masks for objects with diverse spatial scales, MCAE incorporates a multi-scale mask generation strategy, as described in Sec.~\ref{Multi-scale-mask-gen}. 
Specifically, object masks are generated at both the 0.3-meter and 0.6-meter scales, followed by a fusion process that integrates their complementary strengths.
To evaluate its effectiveness, we compare the results of single-scale and multi-scale mask generation, as reported in Tab.~\ref{tab:ablation-multiscale-mask-gen}. 
The 0.3-meter scale produces a substantially larger number of small masks, effectively capturing fine-grained features. 
It proves particularly effective for categories such as \textit{Building} and \textit{Road}, which are often composed of small structures and narrow paths—especially prevalent in rural areas like Fengxian. 
In contrast, the 0.6-meter scale generates fewer but substantially larger masks, which are more effective for capturing broad, homogeneous regions such as \textit{Tree} and \textit{Water}.
The fusion of masks from both scales yields a more balanced distribution of object sizes and substantially increases mask coverage. 
It leads to notable improvements in both mean IoU and OA, demonstrating that combining fine-scale detail with coarse-scale context is critical for achieving high-quality and robust segmentation performance.

\begin{figure*}[!t]
    \centering
    \includegraphics[width=0.9\textwidth]{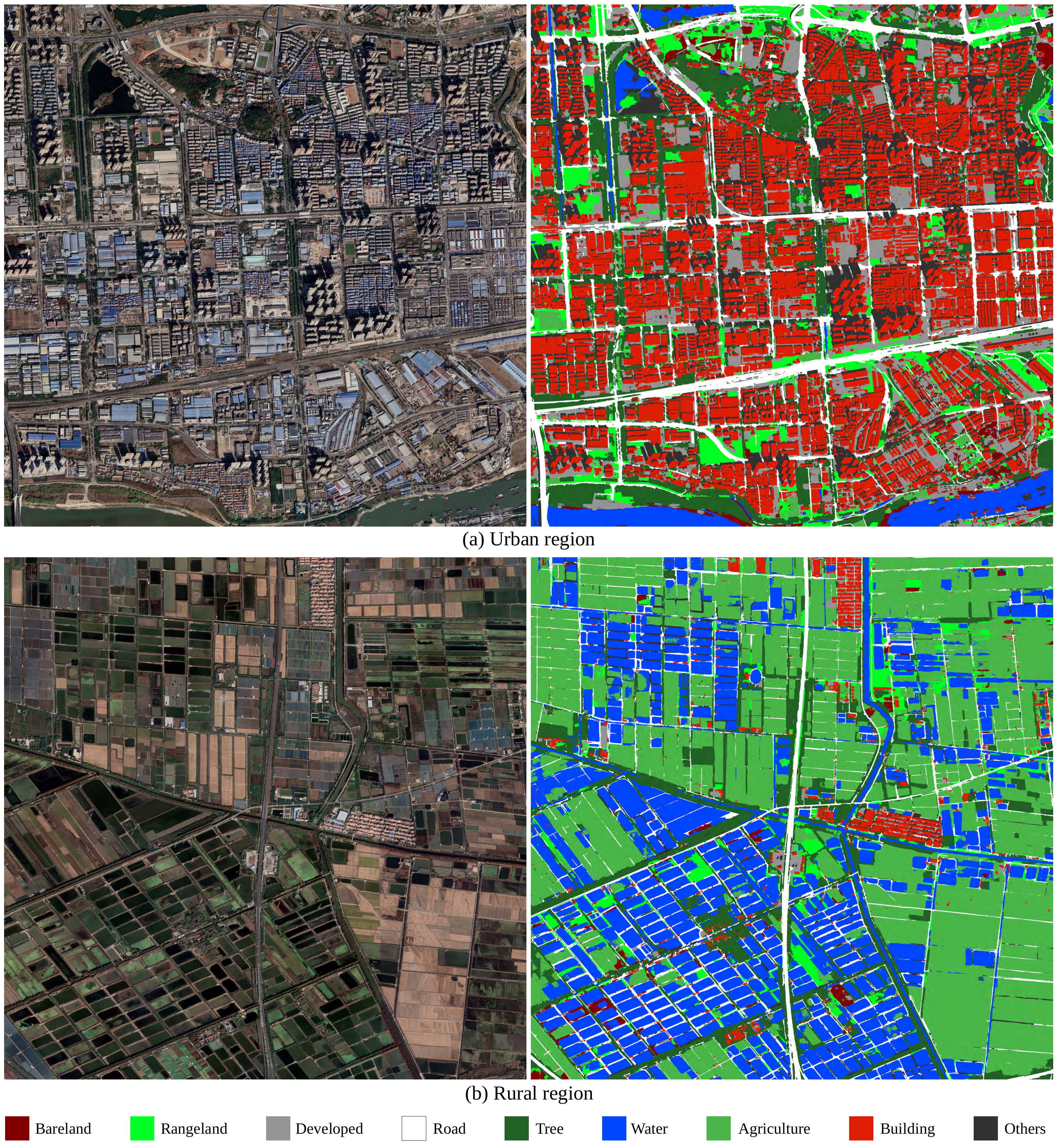}
    \caption{
        Classification results of representative urban and rural regions in Dongxihu District, Wuhan.
    }
    \label{fig:disp-wuhan-dongxihu}   
    \vspace{-0.4cm}
\end{figure*}

\noindent{\textbf{Mask-level feature learning}}.
To enhance the semantic discriminability of mask representations, we adopt a self-supervised learning strategy to optimize mask-level feature extraction, thereby improving the quality of clustering and enabling more accurate and efficient large-scale annotation.
To validate its effectiveness, we compare the results of hierarchical mask clustering using features learned with and without our self-supervised approach, as shown in Tab.~\ref{tab:ablation-stat-mask-level-self-sup}.
We observe a substantial reduction in both the number of clusters and the total number of masks when self-supervised learning is not employed. It is primarily due to the reduced capacity of the features to capture fine-grained inter-class distinctions, thereby compromising the semantic coherence of clusters.
As a result, more low-purity clusters are filtered out to ensure high-quality annotation, leading to a significant decrease in the final annotation volume.

\section{Discussion} \label{Discussion}

\subsection{Comparison with existing datasets}

\textcolor{black}{Based on the proposed MCAE}, we construct HiCity-LC, the first large-scale submeter land cover classification dataset tailored for city-level mapping.
In comparison to existing datasets (Tab.~\ref{tab:comp-lc-dataset}), HiCity-LC offers significant advances in dataset scale, \textcolor{black}{annotation} detail, and geographic coverage. It contains roughly eight times more images than GID~\cite{tong2020land} and Five-Billion-Pixels~\cite{tong2023enabling}, and approximately three times the number of labeled pixels. Compared to OpenEarthMap~\cite{xia2023openearthmap}, HiCity-LC includes an order of magnitude more images and nearly three times the annotation volume, covering a total area of approximately 55,000 square kilometers. Furthermore, it provides fine-grained masks for small-scale landscape elements—such as pedestrian paths within green spaces and boundary ridges between adjacent agricultural plots—that are rarely delineated in existing datasets. Collectively, these characteristics position HiCity-LC as a comprehensive and valuable benchmark that provides critical support for the training and evaluation of foundation models in remote sensing, particularly for pixel-level understanding across diverse and realistic geospatial environments.

\subsection{Generalizability across geographic regions}

The dataset HiCity-LC encompasses five major Chinese cities, Beijing, Tianjin, Shanghai, Chengdu, and Guangzhou, which are strategically chosen to represent the northern, eastern, southern, and western regions. These cities cover diverse climatic zones, vegetation, and urban morphologies. We include full administrative boundaries rather than just urban cores, ensuring representation across high-density centers, suburban zones, rural settlements, and natural areas. The imagery, obtained from Google Earth, incorporates seasonal and lighting variations, further enhancing visual diversity and contributing to model robustness.

To assess spatial generalizability, we apply the model trained on HiCity-LC to the Dongxihu District in Wuhan, a major central city excluded from the training set. As illustrated in Fig.~\ref{fig:disp-wuhan-dongxihu}, the model achieves accurate classification results across both urban and rural settings, indicating strong transferability to previously unseen geographic regions.

\subsection{Practical utility in city-level cropland estimation}
To evaluate the practical utility of our MCAE-based land cover maps, we conduct city-level cropland area estimation using official statistics from municipal bureaus (2020–2024) as a reference. As summarized in Tab.~\ref{tab:cropland-stats}, our estimates show markedly higher consistency with official records compared to SinoLC-1 and WorldCover, particularly in Beijing and Tianjin, where the latter exhibit substantial overestimation. We also observe a strong correlation between SinoLC-1 and WorldCover, which can be attributed to the use of WorldCover as a weak supervision source in the training of SinoLC-1. In contrast, our MCAE-based approach, grounded in high-resolution imagery and fine-grained segmentation, achieves higher precision and yields more actionable cropland statistics.

\begin{table}[!t]
  \centering
  \caption{Estimated cropland areas for five cities based on the land cover maps. \textbf{Reference} are obtained from the official websites of the Municipal Bureaus of Planning and Natural Resources of each city, with area units in hectares.}
  \setlength{\tabcolsep}{3pt}
  \resizebox{\linewidth}{!}{
    \begin{tabular}{lrrrrr}
    \toprule
    \textbf{Name} & \textbf{Beijing} & \textbf{Tianjin} & \textbf{Shanghai} & \textbf{Guangzhou} & \textbf{Chengdu} \\
    \midrule
    Reference & 93,548 & 329,562 & 161,978 & 39,872 & 302,900 \\
    SinoLC-1 & 306,241 & 625,398 & 285,993 & 62,582 & 364,793 \\
    WorldCover & 274,735 & 584,355 & 240,088 & 68,093 & 352,366 \\
    MCAE (ours) & 115,769 & 410,152 & 164,269 & 45,288 & 272,366 \\
    \bottomrule
    \end{tabular}
  }
  \label{tab:cropland-stats}%
\end{table}

\section{Conclusions} \label{Conclusions}

In this study, we present a large-scale dataset tailored for submeter land cover mapping, substantially outperforming existing remote sensing semantic segmentation datasets in terms of scale and diversity. This dataset is constructed using our proposed cost-effective and efficient annotation framework, MCAE. By introducing mask clusters as the fundamental unit of annotation, MCAE integrates multi-scale mask generation, mask-level feature learning, hierarchical mask clustering, and iterative test set curation into an efficient and unified pipeline. It enables the creation of high-quality, semantically diverse, and spatially representative annotations while reducing manual annotation effort by 1-2 orders of magnitude. Extensive experiments conducted in five representative Chinese cities demonstrate that MCAE supports land cover classification with overall accuracies exceeding 85\%. These results suggest that MCAE offers a promising and scalable solution for high-resolution, high-coverage, and cost-efficient semantic mapping on a large scale.

% \section*{Conflict of Interest}


\bibliography{source}
\bibliographystyle{IEEEtran}

\end{document}